# Computation of Total Kidney Volume from CT images in Autosomal Dominant Polycystic Kidney Disease using Multi-Task 3D Convolutional Neural Networks

Deepak Keshwani, Yoshiro Kitamura, Yuanzhong Li

Imaging Technology Center, Fujifilm Corporation, Japan
deepak.keshwani@fujifilm.com

**Abstract.** Autosomal dominant polycystic kidney disease (ADPKD) characterized by progressive growth of renal cysts is the most prevalent and potentially lethal monogenic renal disease, affecting one in every 500–1000 people. Total Kidney Volume (TKV) and its growth computed from Computed Tomography images has been accepted as an essential prognostic marker for renal function loss. Due to large variation in shape and size of kidney in ADPKD, existing methods to compute TKV (i.e. to segment ADKP) including those based on 2D convolutional neural networks are not accurate enough to be directly useful in clinical practice. In this work, we propose multi-task 3D Convolutional Neural Networks to segment ADPK and achieve a mean DICE score of 0.95 and mean absolute percentage TKV error of 3.86%. Additionally, to solve the challenge of class imbalance, we propose to simply bootstrap cross entropy loss and compare results with recently prevalent dice loss in medical image segmentation community.

## 1 Introduction

Autosomal dominant polycystic kidney disease is a hereditary systemic disorder which is characterized by progressive development and growth of bilateral renal cysts filled with fluid [1]. ADPKD effected kidneys can grow as much as 10-15 times in size before complete renal function is lost. It is one of the leading causes of end-stage renal diseases resulting in dialysis or kidney transplantation in majority of the patients. In United States alone, number of patients effected by ADPKD is estimated to be 500,000 [3]. Recently, drugs based on new compound Tolvaptan can slow the rate of cyst growth in ADPKD patients [2]. As per the guidelines, use of the drug is recommended after evaluating the age, stage of ADPKD and whether the disease is progressing rapidly. Rapid progression is defined by total kidney volume (TKV) increase of over 5% per year, where TKV is the combined volume of both left and right kidney. It means that TKV should be measured within 5% precision from Computed Tomography (CT) or Magnetic Resonance (MR) images to be useful in clinical practice. In this work, we target computing TKV from CT images but the methodology is extendable to MR images as well.

Automatic segmentation of ADPK is very challenging due to large changes in its size, shape and position in the abdomen. Fig. 1 shows the difference between contrast enhanced normal kidney and ADPK. A normal kidney roughly measures few hundred milliliters each while an ADPK can measure anywhere from several hundred up to several thousand milliliters. ADPK lose renal function disabling them from filtering contrast agents inserted in blood stream. Thus, kidneys in most ADPKD patients is non-contrast enhanced making the segmentation task even more challenging. Finally, many ADPKD patients develop hepatic cysts which come in contact with renal cysts as shown in Fig. 1. Distinguishing them is the most challenging part in ADPK segmentation, difficult even for a human observer.

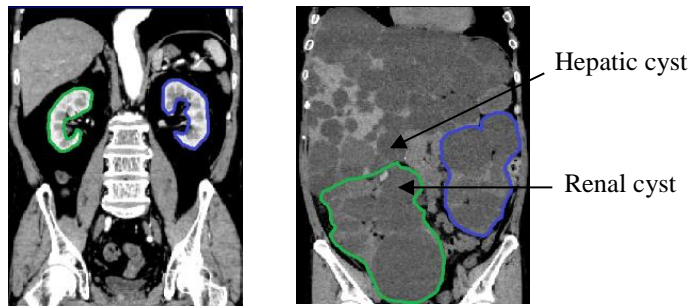

**Fig. 1.** Comparison of normal kidney and ADPK. Left: Contrast enhanced normal kidney, right: non-contrast enhanced ADPK.

A recently proposed automated method to segment ADPK from CT images use 2D fully convolutional neural networks and reports a mean dice score of 0.86 and mean absolute TKV error of over 10% [4]. But as already mentioned that even 5% TKV change can be clinically important. In this work, we improve the ADPK segmentation accuracy to an extent that it can be directly useful for clinical practice with following contributions:

1) We propose a multi-task 3D Fully Convolutional Neural Network (FCN) for ADPK segmentation. The multi-task approach utilizes not only ADPK dataset (abdomen CT images and kidney mask pairs) but also Liver dataset (abdomen CT images and liver mask pairs) for learning ADPK segmentation task. This approach not only surpass existing 2D FCN approach [4] by large margin but also shows improvement over single task 3D FCN approach which learns only from ADPK dataset. Learning from multiple segmentation datasets to boost performance is an interesting solution to data scarcity in medical imaging domain.

2) We propose that by simply bootstrapping cross entropy loss can the resolve class imbalance issue rather than using intricate dice loss recently popular for medical image segmentation tasks.

## 2 Materials and Methods

### 2.1 Dataset and preprocessing

*ADPK dataset:* This dataset is taken from an existing study performed by clinical experts in which they analyze various semi-automated methods to compute TKV from CT images in ADPKD affected patients [1]. The dataset contains a total of 203 abdominal CT image and kidney mask pairs, mostly non-contrast enhanced with various slice thickness ranging from 0.5 to 5mm. Due to small size of the dataset, rather than preparing a separate test set, we perform 3-fold cross validation.

*Liver dataset:* This dataset contains 176 contrast and non-contrast enhanced abdominal CT images with corresponding liver masks, all utilized for training. Most scans in the dataset contain tumorous liver, and only a single image contains hepatic cyst. Note that no two images in Liver and ADKP dataset is of the same patient.

*Preprocessing:* Due to memory limitations, the input images are rescaled to uniform voxel spacing of 1.5 mm. Also, rather than setting an entire abdomen CT image as input to 3D FCN, the images are cropped along z direction (axial) to generate crops roughly of size z =144, y = 250, x = 250. This is true both for the Liver and ADPK dataset. Rotation and scaling is applied as a data augmentation technique to avoid overfitting.

### 2.2 Multi-task 3D Fully Convolutional Network for ADPK segmentation

ADPK shows large variation in its shape, size and position in the abdomen, thus the network should be trained using large and diverse dataset. But, one of the major challenges in medical domain is difficulty in obtaining large datasets validated by clinical experts. This is both due to regulatory hurdles and time consuming 3D labelling task. In our case, the dataset contains 203 3D images and mask pairs, of which a quarter is used for validation. To counter data scarcity challenge, our idea is to use not only ADPK dataset but also datasets of other organs to increase the segmentation accuracy of ADPK. To be specific, we propose a multi-task 3D FCN architecture as shown in Fig. 2 to learn from both the Liver and ADPK dataset.

3D FCN have already shown good performance on anatomy segmentation tasks [5] [6]. Popular 3D FCN architectures like VNet [6] are characterized by a contracting encoder part to extract global features from input image and decoder part to produce full resolution output. In our proposed architecture, encoder layer weights are shared between ADPK and Liver segmentation tasks, with unique decoder part for each task. This is different from recently proposed multi-task multi-modality learning approach using a single encoder-decoder architecture [7]. One of the important reason to split the network at decoder level is due to inconsistent background class definition in the two datasets. In the ADPK dataset, background class encompasses liver region, while in the Liver dataset it encompasses kidney region. Using a single encoder decoder architecture prohibits use of such inconsistent datasets. Fig. 2 illustrates the detailed network

4architecture. When each segmentation task is looked at individually, our network architecture is based on 3D UNet. Although relative to 3D UNet, we increase the depth of our network to have a cumulative receptive field roughly the same size as that of ADPK. In the encoder part, each "conv block" consists of two 3x3x3 convolutional filters each followed by a batch normalization and ReLU unit. The conv block is followed by a 2x2x2 max pooling layer. In the decoder part, each "deconv block" consists of an upconvolution of stride 2 in each dimension. At the end of each decoder network is a 1x1x1 convolution layer to reduce the number of channels to the number of classes in each segmentation task. Finally, we also fuse high resolution feature maps from encoder network to the decoder network using long skip connections, same as 3D UNet architecture. When training, the network is provided with two input images, one from each liver and ADPK dataset.

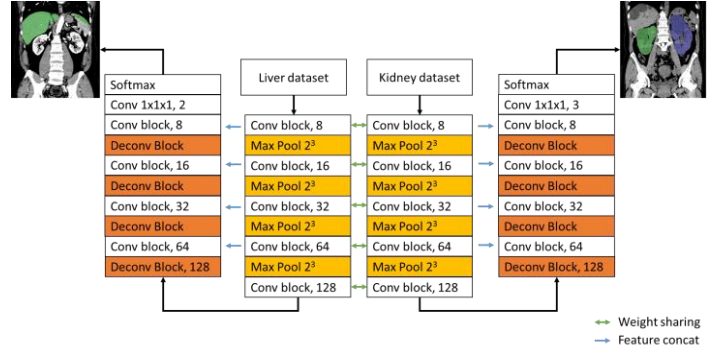

**Fig. 2.** Multi-task 3D FCN: The central towers show shared encoder part of the network whose outputs goes into two separate decoders to generate liver and ADPK segmentation masks.

### 2.3 Bootstrapping cross entropy loss

Class imbalance is a major issue in medical image segmentation, more so when the input is 3D. In this case, the background to kidney voxels ratio is roughly 50:1 which makes conventional cross entropy loss optimization heavily biased towards the background class. To tackle this problem, dice loss has been proposed recently which results in accurate foreground segmentation than simply using weighted cross entropy loss [6]. Let $C$ be the number of classes and let $p_{i,c}$ and $g_{i,c}$ be the prediction and ground truth probabilities belonging to class $c \in \{1, ..., C\}$ at voxel $i \in \{1, ..., N\}$. Then multi-class dice loss is defined in Eq. 1.

$$L_{dice} = -\frac{1}{C} \sum_{c}^{C} \frac{2 * \sum_{i}^{N} p_{i,c} g_{i,c}}{\sum_{i}^{N} p_{i,c} + \sum_{i}^{N} g_{i,c}} \qquad (1).$$

In practice though, multi-class dice loss can be unstable. For example, when input image does not contain kidney region ($\sum_{i}^{N} g_{i,kidney} = 0$), making loss corresponding to kidney class zero ($\sum_{i}^{N} p_{i,kidney} g_{i,kidney} = 0$), even though false positives may exist ($\sum_i^N p_{i,kidney} \neq 0$). Since, we crop the abdomen CT scan image along z dimension, it often happens that the kidney region is not present in the input crop. Additionally, while dice loss does solve the issue of inter-class imbalance to some extent, it does not take into account intra-class imbalance. In this work we propose bootstrapping of cross entropy loss for solving both inter and intra class imbalance following from an existing work in the domain of RGB image segmentation [8]. The idea behind bootstrapping is to backpropagate cross entropy loss not from all but a subset of voxels with posterior probability less than threshold value. Let $y_1, ... y_N \in \{1, ..., C\}$ be the target class labels for voxels $1, .... N$, and let $p_{i,j}$ be the posterior class probability of class $j$ and voxel $i$. Then, the bootstrapped cross entropy loss over $K$ voxels is defined in Eq. 2.

$$L_{bootstrap} = -\frac{1}{K}\sum_{i=1}^{N} 1[p_{i,y_i} < t_K] \log p_{i,y_i} \qquad (2).$$

Where $1[x] = 1$ iff $x$ is true and $t_K \in \mathbb{R}$ is chosen such that $|\{i \in \{1, .... N\}: p_{i,y_i} < t_K\}| = K$. The threshold parameter is determined by sorting the predicted log probabilities and choosing the $K + 1$-th one as the threshold. In this work, we set $K = 0.1N$, meaning that 10% of total voxels participate in the training. Since our problem is a multi-task one, bootstrapped cross entropy loss of both liver and ADKP segmentation tasks are computed separately and then total loss is computed as the mean of two.

## 3    Experiments and training

*2D FCN*: The existing work on ADPK segmentation use 2D FCN on a dataset that is not available publically [4]. So we implement 2D FCN approach as mentioned in the literature on our dataset and report the results. This experiment forms the baseline to which we compare the methods proposed in this work.

*3D FCN:* A standard 3D FCN based on 3D UNet architecture is implemented which learns from ADPK dataset. The architecture is shown in Fig. 2, except that only ADPK part of the network is used for training.

*Multi-task 3D FCN:* We perform multi-task learning using network architecture as shown in Fig. 2 and compare the results with single task learning.

*Multi-task 3D FCN and bootstrapping of cross entropy loss*: Above mentioned 3D FCN and multi-task 3D FCN use dice loss function for optimization. To analyze bootstrapping as a solution to class imbalance, in this experiment we training the multi-task network by minimizing bootstrapped cross entropy loss.

We optimize both dice loss and bootstrapped cross entropy loss using Adam optimizer with a base learning rate of 0.001. All the experiments are run for approximately 100 epochs of ADPK dataset. For multi-task network, one epoch is counted as parsing through entire ADPK and Liver dataset.

## 4 Results

We summarize the mean dice score of left and right kidney in percentage for each experiment in table 2. Valid A, B and C represent validation sets used for 3-fold cross validation.

**Table 2. Comparison of mean left and right kidney dice score in percentage for various experiments**

| Method | Loss | Valid A | Valid B | Valid C |
| --- | --- | --- | --- | --- |
| 2D FCN [4] | Dice | 82.7% | 84.8% | 85.2% |
| 3D FCN | Dice | 94.2% | 93.6% | 94.4% |
| Multi-task 3D FCN | Dice | 94.5% | **94.6%** | **94.8%** |
| Multi-task 3D FCN | Bootstrap cross entropy | **94.9%** | 94.4% | **94.8%** |

Qualitatively, the results are summarized in Fig. 3 where each row represents a different case from ADPK dataset. It is clear that 3D FCN in itself improves the segmentation accuracy by as much as 10% when compared to 2D FCN. 2D FCN produce noisy segmentations (row 1 and 2) because each axial slice is processed independently. Also, the misclassification between liver and kidney in presence of both renal and hepatic cysts is severe as compared to 3D FCN. Although 3D FCN in general resolve such misclassifications (row 1) because it learns from global 3D features, we show cases (row 2) where it fails. In such cases, we find that the proposed multi-task architectures achieve higher accuracy. One explanation could be that the encoder of multi-task architecture generates more rich features than single task architecture, because it is forced to learn an additional task of liver segmentation. Rich features thus make it easier to classify kidney from surrounding organs. The accuracy is also improved in cases where conventional 3D FCN misclassify spleen and kidney (row 3) or when the kidney is contrast enhanced (row 4). Conventional 3D FCN results in poor accuracy in case of high contrast kidneys because of lack of such cases in ADPK dataset. Obtaining better results with multi-task architecture without adding additional contrast enhanced ADPK in the dataset has interesting implications. Thus in summary, while the mean accuracy change from single-task to multi-task might not look significant (table 2), for specific cases we find considerable improvement in accuracy. With respect to loss function; we find that simply bootstrapping cross entropy loss works as good as dice loss or even marginally better (table 2). Results in Fig. 3 show both cases when bootstrap loss performs better (row 1) and vice versa (row 4). With respect to convergence speed, we find no difference between the two as shown in Fig. 4.

We evaluate the clinical usefulness of proposed method using a scatter plot (Fig. 5) which shows TKV error in percentage on entire dataset. The mean absolute error is 3.86% which is below the precision requirement of 5% for clinical applications. Although, for unique test cases or small kidneys, we find multiple cases with absolute error to be larger than 5%. Such cases can be reduced either by increasing the dataset size of ADPK or including additional tasks such as spleen or colon segmentation using multi-task approach.

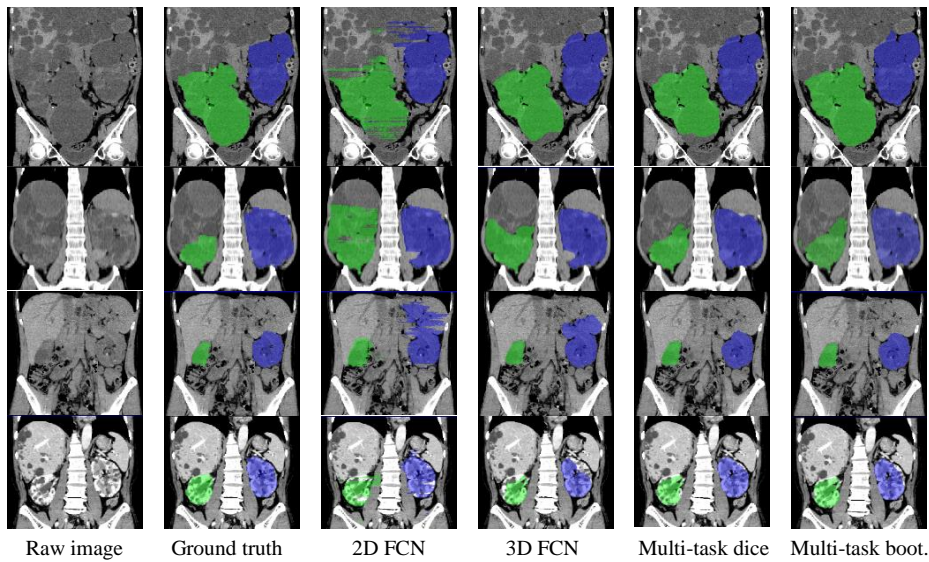

|  |  |  |  |  |  |
|---|---|---|---|---|---|
| Raw image | Ground truth | 2D FCN | 3D FCN | Multi-task dice | Multi-task boot. |

**Fig. 3.** Qualitative comparison of results. Rows represents coronal slice each taken from a different case in ADPK dataset. Columns represent results obtained from different experiments.

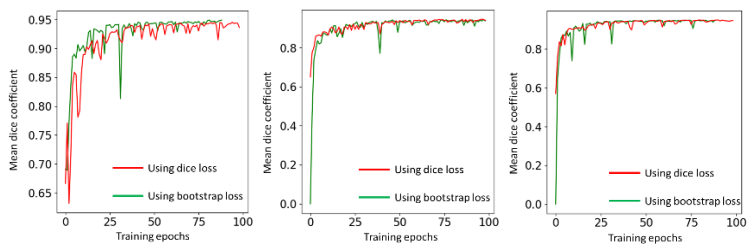

**Fig. 4.** Comparison of mean kidney dice coefficient on validation set A, B and C obtained using multi task dice (Red) and multi-task bootstrap (Green) methodology

## 5    Conclusion and future work

In this work, we propose a multi-task 3D FCN for ADPK segmentation and achieve a mean TKV error acceptable to be directly used in clinical applications. After analyzing both conventional single task 3D FCN approach and proposed multi-task 3D FCN approach we find that multi-task approach improves the segmentation accuracy especially for high contrast ADPK and cases where both renal and hepatic cysts are present. Higher accuracy achieved using multi-task architecture implies that segmentation performance can be improved without explicitly adding data corresponding to target segmentation anatomy. In the future, we would like to analyze the performance by adding additional tasks such as spleen, colon or even segmentation of unrelated anatomies like heart and lung. We also analyzed in this work that simply bootstrapping cross entropy

48

loss works similar to dice loss to counter class imbalance issue. It would be interesting to analyze its performances for even severe inter and intra class imbalance tasks such as aorta segmentation from CT images.

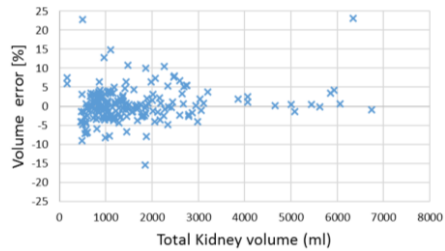

**Fig. 5.** Scattered plot showing percentage TKV error over the entire dataset

## References

1. Muto, S., Kawano, H., Isotani, S., Ide, H., & Horie, S. Novel semi-automated kidney volume measurements in autosomal dominant polycystic kidney disease. Clinical and experimental nephrology, 22(3), 583–590 (2017)
2. Gansevoort, R. T., Arici, M., Benzing, T. Recommendations for the use of tolvaptan in autosomal dominant polycystic kidney disease: a position statement on behalf of the ERA-EDTA Working Groups on Inherited Kidney Disorders and European Renal Best Practice. Nephrology Dialysis Transplantation, 31(3), 337-348 (2016)
3. NIH homepage: https://ghr.nlm.nih.gov/condition/polycystic-kidney-disease#statistics, last accessed 2017/03/02
4. Sharma, K., Rupprecht, C., Caroli, A. Automatic segmentation of kidneys using deep learning for total kidney volume quantification in autosomal dominant polycystic kidney disease. Scientific reports, 7 (1), 2049 (2017)
5. Çiçek, Ö., Abdulkadir, A., Lienkamp, S. S., Brox, T., Ronneberger, O. 3D U-Net: learning dense volumetric segmentation from sparse annotation. Medical Image Computing and Computer-Assisted Intervention 2016, LNCS, volume 9901, pp. 424-432, Springer, Cham (2016)
6. Milletari, F., Navab, N. and Ahmadi, S.A. V-net: Fully convolutional neural networks for volumetric medical image segmentation. In: Fourth International Conference on 3D Vision (3DV), 2016, pp. 565-571, IEEE (2016)
7. Moeskops, P., Wolterink, J.M., van der Velden, B.H., Deep learning for multi-task medical image segmentation in multiple modalities. In: Medical Image Computing and Computer-Assisted Intervention 2016, pp. 478-486, Springer, Cham, (2016)
8. Pohlen, Tobias, Alexander Hermans, Markus Mathias, and Bastian Leibe. Full-resolution residual networks for semantic segmentation in street scenes. Proceedings of the IEEE conference on Computer Vision and Pattern Recognition, pp. 4151-4160 (2017)

## Acknowledgements
We acknowledge using Reedbush-L (SGI Rackable C2112-4GP3/C1102-GP8) HPC system in the Information Technology Center, The University of Tokyo for GPU computational resources used in this work.